\documentclass[preprint,12pt]{elsarticle} 
\usepackage[numbers]{natbib}

\usepackage[utf8]{inputenc}
\usepackage[english]{babel}
\usepackage{csquotes}
\usepackage{algorithm}
\usepackage[noend]{algpseudocode}
\algrenewcommand\algorithmicrequire{\textbf{Input:}}
\algrenewcommand\algorithmicensure{\textbf{Output:}}
\algnewcommand{\algorithmicand}{\textbf{ and }}
\algnewcommand{\algorithmicor}{\textbf{ or }}
\algnewcommand{\OR}{\algorithmicor}
\algnewcommand{\AND}{\algorithmicand}

\usepackage{amsmath}
\usepackage{amsfonts}
\usepackage{amssymb}
\usepackage{amsthm}
\usepackage{nicefrac}
\usepackage{algpseudocode,algorithm,algorithmicx}
\usepackage[short]{optidef}
\usepackage{mathrsfs}
\usepackage{colonequals}
\usepackage{diagbox}

\usepackage[hidelinks]{hyperref}
\hypersetup{colorlinks=true, urlcolor=blue}

\makeatletter
\newcommand*{\defeq}{\mathrel{\rlap{%
                     \raisebox{0.3ex}{$\m@th\cdot$}}%
                     \raisebox{-0.3ex}{$\m@th\cdot$}}%
                     =}
\makeatother

\usepackage{tabularx}
\usepackage{makecell}

\usepackage{pgfplots}
\usepackage{pgfplotstable}
\usepackage{tikz}
\usetikzlibrary{arrows, positioning, calc}
\usetikzlibrary{automata,snakes}
\usetikzlibrary{decorations}
\usetikzlibrary{decorations.markings,decorations.pathreplacing}
\usetikzlibrary{shapes.misc}
\usetikzlibrary{backgrounds, intersections}

\pgfplotsset{compat=1.14} 

\usepackage{amssymb}

\usepackage{todonotes} 

\journal{}

\begin{document}

\begin{frontmatter}



\title{A machine learning based algorithm selection method to solve the minimum cost flow problem}

\renewcommand*{\thefootnote}{\fnsymbol{footnote}}
\author[a]{Philipp Herrmann}
\ead{pherrman@rhrk.uni-kl.de}
\author[a]{Anna Meyer}
\ead{meyera@rhrk.uni-kl.de}
\author[a]{Stefan Ruzika}
\ead{ruzika@mathematik.uni-kl.de}
\author[b]{Luca E. Schäfer\corref{mycorrespondingauthor}}
\cortext[mycorrespondingauthor]{Corresponding author}
\ead{lucaelias.schaefer@aol.com}
\author[a]{Fabian von der Warth}
\ead{vdwarth@rhrk.uni-kl.de}

\address[a]{Department of Mathematics, Technische Universität Kaiserslautern, 67663 Kaiserslautern, Germany}
\address[b]{Comma Soft AG, 53229 Bonn, Germany}

\begin{abstract}
The minimum cost flow problem is one of the most studied network optimization problems and appears in numerous applications. Some efficient algorithms exist for this problem, which are freely available in the form of libraries or software packages. It is noticeable that none of these solvers is better than the other solution methods on all instances. Thus, the question arises whether the fastest algorithm can be selected for a given instance based on the characteristics of the instance.
To this end, we train several machine learning classifiers to predict the fastest among a given set of solvers. We accomplish this by creating a representative data set of 81,000 instances and characterizing each of these instances by a vector of relevant features. To achieve better performance, we conduct a grid search to optimize the hyperparameters of the classifiers. Finally, we evaluate the different classifiers by means of accuracy. It is shown that tree-based models appear to adapt and exploit the relevant structures of the minimum-cost flow problem particularly well on a large number of instances, predicting the fastest solver with an accuracy of more than  90\%.
\end{abstract}




\end{frontmatter}

\section{Introduction}\label{sec:intro}
\paragraph{Motivation}

The minimum cost flow (MCF) problem is one of the most classical problems in network optimization: Given a directed graph with capacitated arcs, a linear cost function on the arcs as well as a set of sources and sinks with supply/demand values, the goal is to compute a feasible flow having minimal cost. Here, feasible refers to the fact that the flow obeys the (lower and upper) capacity bounds on each arc and satisfies flow conservation in each vertex.

The MCF problem and its variants are highly relevant in many applications \cite{mcfsurvey}, for example, in evacuation planning \cite{mcf_application_evacuation_modeling_1,mcf_application_evacuation_modeling_2,mcf_application_evacuation_modeling_3,mcf_application_evacuation_modeling_4,mcf_application_evacuation_modeling_5}, freight traffic \cite{mcf_application_freight_traffic_1,mcf_application_freight_traffic_4,mcf_application_freight_traffic_3,mcf_application_freight_traffic_5,mcf_application_freight_traffic_2}, transport logistics \cite{mcf_application_transport_1,mcf_application_transport_3,mcf_application_transport_2}, and many more. 

There exist several algorithmic approaches for solving the MCF problem. The first to develop such an algorithm were Ford and Fulkerson in 1962~\cite{ford_and_fulkerson} by modifying Kuhn's Hungarian Method~\cite{kuhn}.
This algorithm does not run in polynomial time, in constrast to the capacity scaling algorithm proposed by Edmonds and Karp in 1972~\cite{ssp4} which runs in weakly polynomial time. 
The first strongly polynomial time algorithm has been developed by Tardos in 1985~\cite{tardos}. An extensive overview of the history of algorithms for the MCF problem can be found in~\cite{mcfalgorithms}.

Of the many algorithms that have been published on the MCF problem,  we consider in this article seven methods for which freely available implementations exist: The simple cycle canceling algorithm (SCC), the minimum mean cycle canceling algorithm (MMCC) and the cancel and tighten algorithm (CAT) use a variation of cycle canceling in the residual network of a given feasible flow~\cite{MMCC, scc}.
The successive shortest path algorithm (SSP) and capacity scaling algorithm (CAS) are dual algorithms. They maintain an optimality criterion during the execution of the algorithm and strive for feasibility by computing shortest paths from surplus to deficit vertices~\cite{ssp4}. The network simplex algorithm (NS) is a variant of the simplex method specialized for the MCF problem \cite{ns1, ns2}.
Finally, we also consider a cost scaling algorithm (CS2) \cite{cos}. We discuss the details later in Section \ref{sec:algorithms}.

It has been shown in \cite{mcfalgorithms} that the performance of MCF algorithms highly depends on the instance.
For example, it has been observed that the network simplex algorithm performs well on smaller instances, while, on larger instances, it often performs worse than cost scaling approaches.
Or, there is a tendency that dual algorithms based on the successive computation of shortest paths often perform better than other algorithms on road network instances but are generally worse than other algorithms on grid-like instances. On smaller grid-like instances sometimes cost-scaling algorithms are the fastest. To conclude, these observations are subject to some variations depending on the particular instance and rules of thumb derived from them are inaccurate.

In order to minimize the total running time of a system which is based on solving the MCF problem, it seems a natural idea to select an algorithm depending on a given MCF instance. We explore this idea by testing several machine learning (ML) based selection algorithms. A similar idea has been pursued for the traveling salesperson problem~\cite{tspfeature} and for the graph coloring problem~\cite{gcpfeature}.  In both cases, the authors show that using a set of problem-specific features, the accuracy of selecting the algorithm using those methods improves over the selection of the algorithm which performs best on average. We follow the same approach in this paper.

\paragraph{Outline}
The remainder of this paper is organized as follows. 
In Section~\ref{sec:notation}, the basic notation and problem setting is introduced. In Section~\ref{sec:algorithms}, a brief introduction to the min cost flow algorithms considered is given. Section~\ref{sec:method} then goes into detail about the selection method developed and, in Section~\ref{sec:results}, the results are presented and discussed.
Finally, Section~\ref{sec:conclusion} concludes the paper.

\paragraph{Source code} The presented results as well as the trained models and evaluations can be found at the following repository: \href{https://github.com/fabianvdW/mcf_alg_selection}{github.com/fabianvdW}

\section{Preliminaries}\label{sec:notation}
Let $G=(V,A)$ be a directed graph with vertex set $V$ and arc set $A$. By $n\defeq \lvert V \rvert$ and $m \defeq \lvert A \rvert$, we denote the number of vertices and arcs, respectively.
Further, by $\delta^+(i)$ and $\delta^-(i)$, we denote the set of all outgoing and incoming arcs, respectively.
In the MCF problem, each arc~$(i,j) \in A$ has a cost value~$c_{i j} \in \mathbb{Z}_{\geq0}$ for sending one unit of flow along the arc and an upper capacity value $u_{i j}\in \mathbb{Z}_{\geq0}$ restricting the maximum amount of flow which can be sent along arc~$(i,j)$. Additionally, each vertex~$i\in V$ is associated with a supply value~$b_i \in \mathbb{Z}$. If $b_i > 0$, then vertex~$i$ is called a source. If $b_i < 0$, then vertex~$i$ is called a sink. The quadruple~$(G, c, u, b)$ is called an MCF instance. In the MCF problem, the task is to send flow along the arcs of the graph such that the excess value of each vertex corresponds exactly to its supply value, while minimizing the overall costs and not violating the upper bound constraints. This can also be expressed as an optimization problem (cf.~Problem~\ref{eq:mcf}) in the following way.
\begin{mini!}
    {}{\sum_{(i,j) \in A} c_{i j}x_{i j}}
    {\label{eq:mcf}}{}
    \addConstraint{\sum_{(i,j)\in \delta^+(v)} x_{i j}- \sum_{(i,j)\in \delta^-(v)} x_{i j}}{= b_v, \quad \label{eq:conservation}}{v\in V}
    \addConstraint{0\leq x_{i j}}{\leq u_{i j},\quad \label{eq:capacity}}{(i,j)\in A.}
\end{mini!}
By $x$, we denote a solution to Problem~\ref{eq:mcf} satisfying constraints \eqref{eq:conservation} and \eqref{eq:capacity}, which are known as  flow conservations constraints and flow capacity constraints, respectively. An MCF instance is called feasible, if there exists a solution.

The concept of a flow can be generalized to so-called pseudoflows, which may not satisfy flow conservation. More precisely, the excess of a vertex~$v$ is defined as $$e_v = b_v + \sum_{(i,j)\in \delta^-(v)} x_{i j} - \sum_{(i,j)\in \delta^+(v)} x_{i j},$$ where $x$ satisfies \eqref{eq:capacity}. Thus, for a feasible flow, the excess of every vertex is equal to zero. If, for a given function $x:A \rightarrow \mathbb R$, the excess of vertex $v$ is not necessarily zero but $x_{ij} \leq u_{ij}$ for all arcs~$(i,j)$, then $x$ is called a \textit{pseudoflow}.

The feasibility of an instance already implies that $\sum_{i\in V}b_i =0$, cf. \cite{mcfsurvey}. We denote the total supply by $\sum_{i \in V: b_i >0}b_i$.

Let $x$ be a feasible flow. Then, the corresponding \textit{residual network} $G_x = (V, A_x)$ with respect to $x$ is defined as follows. For each arc $(i,j) \in A$, there is a \textit{forward arc} $(i,j) \in A_x$ with capacity $r_{ij} = u_{ij} - x_{ij}$ and cost $c_{ij}$ if $r_{ij} > 0$. Further, there exists a \textit{backward arc} $(j,i) \in A_x$ with capacity $r_{ji} = x_{ij}$ and cost $-c_{ij}$, if $x_{ij} > 0.$

Further, $x$ is called $\epsilon$-optimal for a given $\epsilon\geq0$ if there exists a vertex potential function $\pi$ such that $c_{ij}^\pi \geq -\epsilon$ for all arcs $(i,j)$ in the residual network, where $c_{ij}^\pi = c_{ij} + \pi_i - \pi_j$. It can be shown that, if $\epsilon < \frac{1}{n}$, then an $\epsilon$-optimal feasible flow is optimal \cite{theory}.

\section{Minimum cost flow algorithms}\label{sec:algorithms}

In this section, we briefly review state-of-the-art algorithms for solving the minimum cost flow problem which are considered in this article. For a more detailed overview see~\cite{mcfsurvey}. We consider seven algorithms, six of which are provided by the Lemon library, which is in line with the approach taken in \cite{mcfalgorithms}. Lemon is an abbreviation of \textit{Library for Efficient Modeling and Optimization in Networks}. It is an open source C++ library that focuses on combinatorial optimization problems on graphs and networks. The library provides several efficient implementations of graph algorithms such as shortest path, spanning tree and minimum cost flow algorithms. For more information on Lemon see \cite{lemon}.

The specific cost scaling algorithm considered in this article is not available in Lemon. Instead, we use an implementation by Goldberg and Cherkassky in C language. It is an authoritative implementation, which is seen as a benchmark for solving the MCF problem, see~\cite{mcfalgorithms2,cs2}. This implementation turned out to be slightly superior than the corresponding Lemon implementation according to the studies of Kovács in \cite{mcfalgorithms}. For more details on the Goldberg-Cherkassky implementation, we refer to~\cite{cs2}.

\subsection{The simple cycle canceling algorithm - SCC}\label{subsec:scc}
This algorithm has originally been proposed by Klein in 1967~\cite{scc}. First, a feasible solution is computed, for example by means of a maximum flow computation~\cite{theory}. The algorithm then maintains feasibility and iteratively reduces the cost of the current flow until no improvement is possible. To improve the cost of the flow, a cycle with negative cost is found in the current residual network and the flow is augmented along this cycle. The algorithm terminates if the residual network does not contain any negative cycle. This implies the optimality of the solution~\cite{theory}. The implementation provided by Lemon applies the Bellman-Ford algorithm, see \cite{bellman1958routing}, to find negative cycles.

\subsection{The minimum mean cycle canceling algorithm - MMCC}\label{subsec:MMCC}
Goldberg and Tarjan have introduced a special variant of the aforementioned SCC algorithm in 1988~\cite{MMCC} which only differs in the selection criterion of the negative cycle in the residual network: Instead of choosing an arbitrary cycle with negative cost in each iteration, a negative cycle with minimum mean cost is chosen.

\subsection{The cancel and tighten algorithm - CAT}\label{subsec:CAT}
This variant of a cycle cancelling algorithm has also been introduced by Goldberg and Tarjan~\cite{MMCC}. It ensures $\epsilon$-optimality of the solution in each iteration and decreases the $\epsilon$ values during the execution of the algorithm by performing two main steps at each iteration, i.e., the \textit{cancel} step and the \textit{tighten} step. In the \textit{cancel} step, the cycles, which consist entirely of arcs with negative reduced costs, get canceled by sending as much flow as possible along these cycles. In the \textit{tighten} step, the vertex potentials are updated by means of minimum mean cost cycle computations. This decreases $\epsilon$ to at most $(1 - \frac{1}{n})$ times its former value, cf.~\cite{MMCC}.

\subsection{The successive shortest path algorithm - SSP}\label{subsec:SSP}
The SSP has been established in 1972 by Edmonds and Karp~\cite{ssp4}. However, a former versions that do not use vertex potentials have already been developed by Jewell~\cite{ssp2} (1958), Iri~\cite{ssp1} (1960) and Busacker and Gowen~\cite{ssp3} (1961). 

The algorithm starts with the zero flow, which is a pseudoflow and has an objective value of zero. This pseudoflow satisfies the reduced cost optimality criterion, however, it is in general not a feasible flow~\eqref{eq:conservation}. Thus, the algorithm strives to achieve feasibility while maintaining optimality. 

More specifically, in each iteration, a pair of vertices $u,v$ with excesses $e_u > 0$ and $e_v < 0$ is selected. A shortest path from $u$ to $v$ in the residual network with respect to the reduced arc costs is computed. The current solution is augmented along this path and the vertex potentials are increased by the shortest path distance. This maintains the non-negativity of the reduced arc costs in the residual network, which in turn ensures the optimality of the current pseudoflow, cf.~\cite{theory}. If $e_v = 0$ for all vertices $v\in V$, the algorithm terminates, since the flow is then feasible. The Lemon implementation uses Dijkstra's algorithm for the shortest path computations.

\subsection{The capacity scaling algorithm - CAS}\label{subsec:CAS}
The CAS, which is also due to Edmonds and Karp \cite{ssp4}, improves the SSP algorithm by augmenting flow by a sufficiently large amount in each iteration. In a $\Delta$-scaling phase, only those vertices with $e_u \geq \Delta$ and $e_v \leq -\Delta$ are considered, where $\Delta>0$ is a sufficiently large power of two. Along the shortest $u-v$ path, the solution is augmented by $\Delta$ units of flow. If no such pair of vertices exists, $\Delta$ is  divided by two and the algorithm continues with the next phase. The algorithm terminates at the end of the phase where $\Delta = 1$ with an optimal solution.

\subsection{The network simplex algorithm - NS}\label{subsec:NS}
This algorithm for solving the MCF problem has been devised by Dantzig in 1951, cf.~\cite{ns1, ns2}.
The idea of this algorithm is based on the primal simplex algorithm for linear programming, cf.~\cite{mcfalgorithms}. The bases are (undirected) spanning trees. The arc set is partitioned into three subsets $T,L$ and $U$ where the arcs in $T$ form a spanning tree with the network flow values fulfilling the capacity constraints \eqref{eq:capacity}. The amount of flow at the arcs in $U$ is fixed at the capacity of the arc while all arcs in $L$ have zero flow. A solution of this form is called a \textit{spanning tree solution}. If an MCF instance has an optimal solution, there exists a spanning tree solution which is optimal, cf.~\cite{mcfalgorithms}.

The algorithm starts with a feasible spanning tree solution and maintains its feasibility throughout the iterations and successively reduces the costs. This is achieved by finding a non-tree arc, i.e., an arc in $L$ or $U$ violating the optimality conditions. This arc is added to the spanning tree, which creates a negative cycle. By augmenting flow along it, the cycle gets canceled and the tree structure is regained. The algorithm terminates, if no entering arc can be found and if the current solution is optimal. 

\subsection{Cost scaling algorithm - CS2}\label{subsec:CS2}
This algorithm has been proposed by Röck in 1980, cf.~\cite{cos}. It takes an $\epsilon$-optimal solution and successively decreases $\epsilon$ by dividing it by a given value of $\alpha > 1$ until $\epsilon < \frac{1}{n}.$ 

At the beginning of each iteration, an $\epsilon$-optimal solution $x$ is turned into an optimal pseudoflow and $\epsilon$ is divided by $\alpha$. An arc $(i,j)$ in the residual network~$G_x$ is called \textit{admissible}, if it has negative reduced costs. Then, a vertex $v$ with $e_v > 0$ is selected. If there exists an admissible arc outgoing  of $v$, the \textit{push}-operation pushes flow along this arc. If no such arc exists, a \textit{relabel}-operation is executed. In this step, the potential of the vertex $v$ is decreased by the largest possible amount to be still $\epsilon$-optimal. This creates new admissible arcs, since it decreases the reduced costs of the outgoing arcs and therefore \textit{push}-operations are again possible. The iteration is finished when the excess of all vertices is zero, which implies the feasibility of the solution.

\section{ML-based algorithm selection method}\label{sec:method}

The \textit{Algorithm Selection Problem}, which has been formulated by J. R. Rice \cite{algselection} in 1976, is about finding the most \enquote{suitable} algorithm for a given problem. More specifically, with regard to our setting, the problem task
is to select the algorithm that performs best on a given MCF instance from the seven algorithms presented in Section~\ref{sec:algorithms}.

In the first step of the ML-based algorithm selection method, we generate a large and representative set of MCF instances. Each of these MCF instances is then characterized by a vector of features that can be efficiently computed. We then evaluate the running times of the aforementioned algorithms on this data, and in turn train different ML algorithms that can predict the fastest algorithm for a given MCF instance.

In the further course of this section we explain the above aspects in more detail. To this end, the remainder of this section is structured as follows. In Subsection~\ref{subsec:data}, we present the data generation process, define the datasets and record the performances of the respective algorithms on the data. Subsequently, in Subsection~\ref{subsec:features}, we provide a precise definition of the vector of features. In Subsection~\ref{subsec:mlalgorithms} and Subsection~\ref{subsec:training}, we present the ML algorithms and the training methodology used, respectively. Afterwards, we discuss our results in Section~\ref{sec:results}.

\subsection{Data generation and datasets}\label{subsec:data}
Multiple random generators are used to generate a representative dataset of MCF instances of various sizes and characteristics. More specifically, we use the generators Netgen, Gridgen, Gridgraph\footnote{Source code at \href{}{http://archive.dimacs.rutgers.edu/pub/netflow/generators/network/}} and Goto. These well-established generators have been used in several previous works, see e.g. \cite{generators}, \cite{mcfalgorithms2}, \cite{cs2} and \cite{mcfalgorithms}.

In the following, we briefly discuss for each generator which set of parameters is used to create the instances. Note that not all generated instances are feasible. This infeasibility issue usually results from bad parameter constellations, such as a high total supply value, with low capacity values at the same time.  These instances are to be sorted out at a later stage once the algorithms are evaluated on them.\\
\textbf{Netgen} uses as parameters the number of vertices $n$ and arcs $m$, respectively, the total positive supply, the number of supply and demand vertices, as well as the maximum arc cost and capacity. The specific parameter values used can be seen in Table \ref{tab:parameters}. In total, we generate 18000 instances, 15268 of which are feasible.\\
\textbf{Gridgen} utilizes basically the same parameters as Netgen. Additionally,the width of the generated grid is specified as well as whether two-way arcs or one-way arcs are to be generated. In the two-way arc option there are two arcs, one in each direction, between every adjacent pair of vertices. With the latter option, only one arc with alternating direction is generated. We use the same range of parameter values as for Netgen, while setting the width either to $\sqrt{n}$ or $\sqrt{n/2}$. Again, we generate 18000 instances, 17980 of which are feasible.

\textbf{Gridgraph}, similar to Netgen, also takes the supply, the maximum costs and capacities as parameters. The number of vertices is implicitly given by the determination of the width and length of the grid. For the width and length, we use 5, 10, 20, 30, 50, 70, and 100 vertices, respectively, while the other parameters remain the same as for Netgen.  We fix the number of supply and demand vertices to one. We generate 27000 Gridgraph instances, 21962 of which are feasible instances.

\textbf{Goto} uses only the number of vertices $n$, the number of arcs $m$, as well as the maximum costs and capacities as parameters, see Table \ref{tab:parameters} for the exact parameter values used. The number of supply and demand vertices is fixed to one and the total supply value is adjusted by the generator depending on the arc capacities. Finally, of the 18000 instances generated, only 80 are infeasible. However, the generator is more restrictive on the parameter values than the other generators. Therefore, we get fewer different parameter combinations, but generate more instances with the same setting.

\renewcommand{\arraystretch}{1.25}
\begin{table}[H]
    \centering
    \begin{tabularx}{\textwidth}{|X|X|X|}
    \hline
        \textbf{Parameter} & \textbf{Netgen} & \textbf{Goto} \\
        \hline
        number of vertices & 64,128,256,512,1024 & 64,128,256,512,1024\\
        \hline
        number of arcs & $8n,n\sqrt[4]{n},n\sqrt{n}$ & $8n,n\sqrt{n}$ \\
        \hline
        total supply & $\sqrt{n}$, 10$\sqrt{n}$, 100$\sqrt{n}$, 1000$\sqrt{n}$ & adjusted by generator \\
        \hline
        supply vertices & 1, $\sqrt[4]{n},\sqrt{n}$ & 1 \\
        \hline
        max arc costs & 2, 10, 100, 1000, 10000 & 10, 100, 1000, 10000\\
        \hline
        max arc capacity & 1, 10, 100, 1000, 10000 & 10, 100, 1000, 10000\\
        \hline
    \end{tabularx}
    \caption{Parameters used for the generators Netgen and Goto. All values are rounded to integers.}
    \label{tab:parameters}
\end{table}
\renewcommand{\arraystretch}{1}

\begin{figure}[H]
 \centering
 \includegraphics[width=12cm,height=9cm,keepaspectratio]{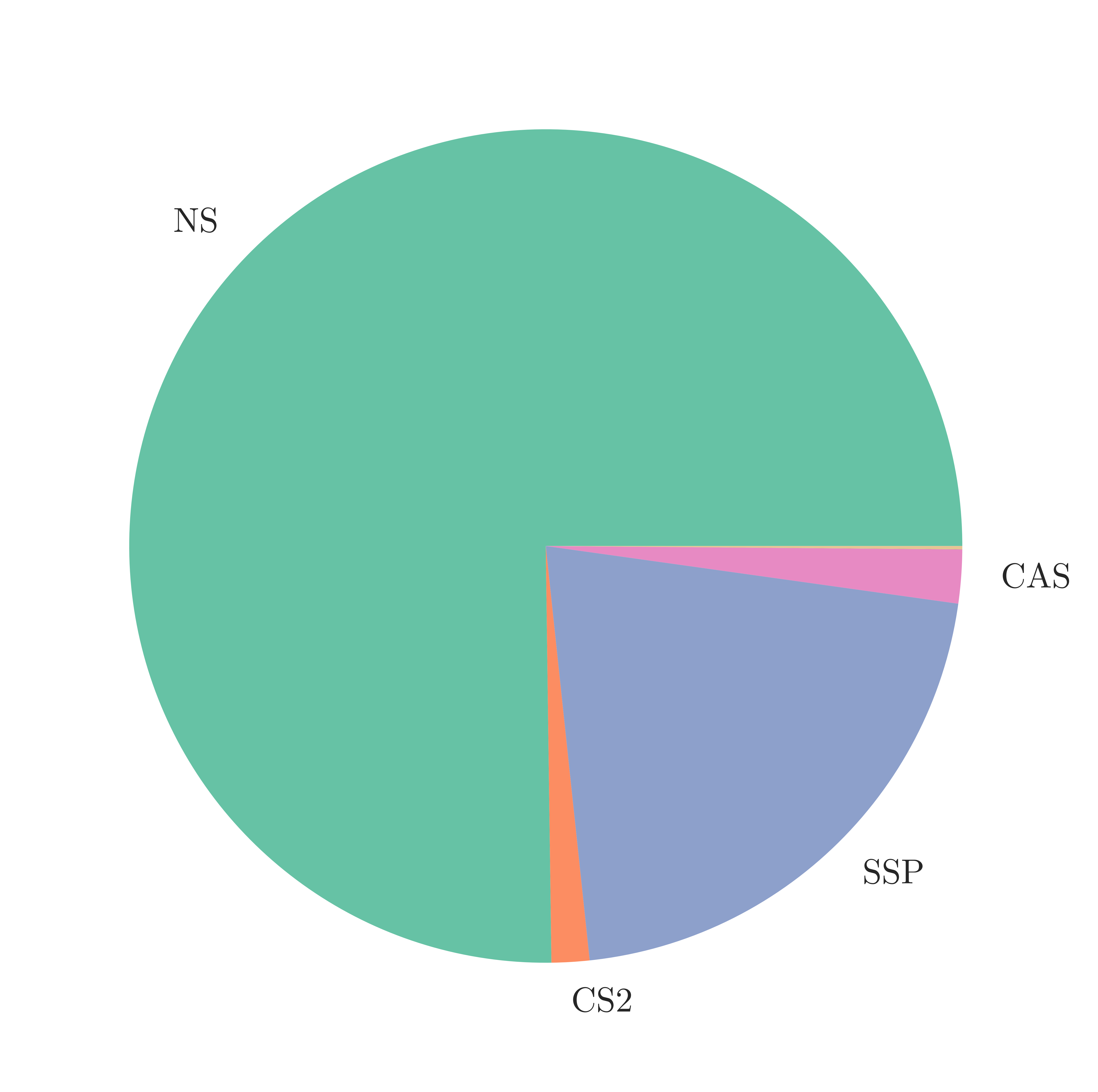}
 \caption{Distribution of the fastest algorithms for all 73130 feasible instances.}
 \label{fig:algos_all}
\end{figure}

In total, we generate 81000 instances, 73130 of which are feasible. We run all seven algorithms, see Section \ref{sec:algorithms}, on these instances and record the respective running times. Figure \ref{fig:algos_all} and Table \ref{tab:algos} show for how many instances each algorithm is the fastest. The Network Simplex algorithm, see Subsection \ref{subsec:NS}, turns out to be the fastest with a rough estimate of $75\%$ of the instances. The second best algorithm is the successive shortest path algorithm, see Subsection \ref{subsec:SSP}, which is the best choice for more than $20\%$ of the instances. The cost scaling algorithm, see Subsection \ref{subsec:CS2}, and the capacity scaling algorithm, see Subsection \ref{subsec:CAS}, are the fastest algorithms for more than 1000 and 1500 instances, respectively, which correspond to $1.5\%$ and $2\%$ of the instances. The cycle canceling algorithms, see Subsection \ref{subsec:scc} to Subsection \ref{subsec:CAT}, however, only play a minor role, as all three algorithms together are the fastest for less than 100 instances. Thus, the corresponding slices in the pie chart are too small to be visible.

\begin{table}[H]
    \centering
    \begin{tabularx}{\textwidth}{|p{1.3cm}|p{2cm}|p{2.3cm}|p{3cm}|p{1.4cm}|p{1.1cm}|}
    \hline
        \textbf{Alg.} & \textbf{Netgen} & \textbf{Gridgen} & \textbf{Gridgraph} & \textbf{Goto} & \textbf{All} \\
        \hline
        SCC  & - & 1 & 5 & - & 6\\
        \hline
        MMCC  & - & - & 2 & - & 2 \\
        \hline
        CAT & - & - & 84 & -  & 84\\
        \hline
        SSP & 2122 & 3477 & 9836 & -  & 15435\\
        \hline
        CAS & 113 & 200 & 1212 & - & 1525\\
        \hline
        NS & 12250 & 14100 & 10743 & 17913 & 55006\\
        \hline
        CS2 & 783 & 202 & 80 & 7 & 1072\\
        \hline
        Total & 15268 & 17980 & 21962 & 17920 & 73130\\
        \hline
    \end{tabularx}
    \caption{Number of instances in which each algorithm is the fastest, with an additional differentiation with respect to the different generators. The last row lists the number of feasible instances.}
    \label{tab:algos}
\end{table}

\begin{figure}[H]
 \centering
 \includegraphics[width=13cm,height=10cm,keepaspectratio]{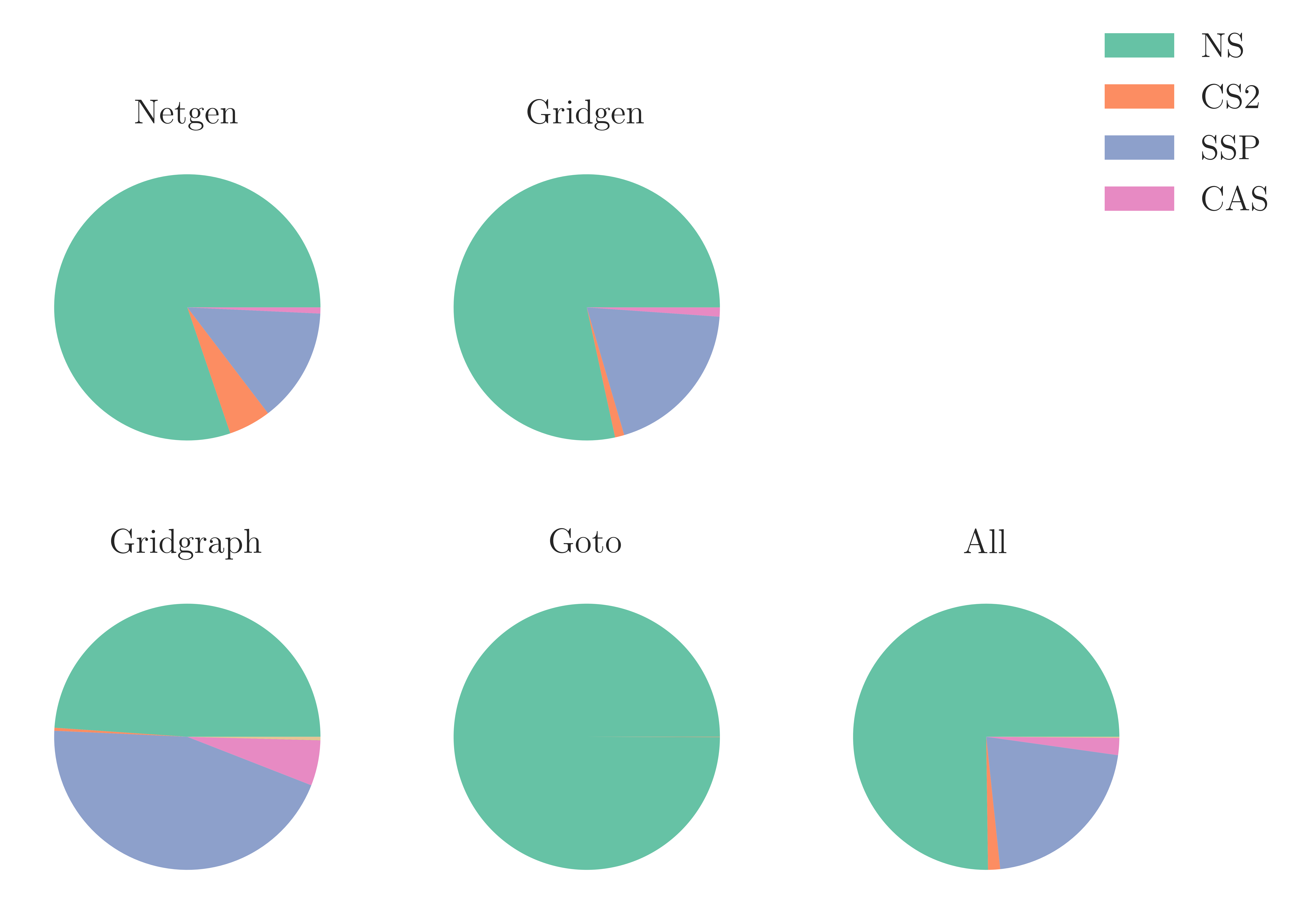}
 \caption{Distribution of the fastest algorithms on 15268 Netgen, 17980 Gridgen, 21962 Gridgraph and 17920 Goto instances and combined on all 73130 feasible instances.}
 \label{fig:algos_generators}
\end{figure}

Figure \ref{fig:algos_generators} shows how often each algorithm most quickly determines the optimal solution with respect to the respective instances of the different generators. The exact numbers can be seen in Table \ref{tab:algos}. On the Gridgen and Netgen instances, the distribution is similar to the distribution on all feasible instances. However, the NS and CS2 percentages are slightly higher than Gridgen at $80\%$ and $5.13\%$, respectively. Thus, the share of the SSP and CAS algorithm drop to less than $15\%$ and $1\%$, respectively. Note that the cycle canceling algorithms play no role at all.

The dominance of the network simplex algorithm is even greater on the Goto instances. The NS is the best choice for all but seven instances, where CS2 is the fastest algorithm.

Regarding the Gridgraph instances, none of the described algorithms dominates as much as on the other three generators, i.e., Gridgen, Netgen, Goto. Nevertheless, the NS algorithm is still the best algorithm for these instances, as it is the fastest for $48.9\%$ of the instances. However, the SSP algorithm is almost as good,  winning just shy of $45\%$ of the instances. CAS is the best choice for more than $5\%$ of the instances. On the Gridgraph instances, CS2 does not perform as well as on the other instances. It is the fastest algorithm only on 80 instances, which is less than $0.4\%$. Conversely, the CAT algorithm, which does not win any Netgen, Gridgen or Goto instance, is the fastest algorithm on a total of 84 of the Gridgraph instances. The other two cycle canceling algorithms, however, only play a minor role on this generator either.

\subsection{Features}\label{subsec:features}
In order to appropriately apply the ML algorithms to be described in Section \ref{subsec:mlalgorithms}, we characterize each MCF instance by a vector of features. Table \ref{tab:features} shows the 21 features that we calculate for each instance. Many of the parameters we use for creating the instances with the generators also appear in the feature vector, since these basic parameters are important for the description of each instance. Thus, the number of vertices and arcs, the total supply as well as the number of supply and demand vertices denote some of our features. Further, we calculate the maximum and minimum of the arc costs and capacities appearing in the graph, as well as their mean value, sum and standard deviation. Additionally, the mean value and the standard deviation are divided by the maximum arc cost and capacity, respectively. Thereby, these features are independent of the input size.

Gridgen and Gridgraph generate graph instances in a so called grid-shape. To this end, we determine the number of arcs between one source vertex and one sink vertex to identify the shape of the grid, for instance if it is rather long or wide. Although some Gridgen instances have more than one source and demand vertex, we specify the amount of arcs only between one source and one sink vertex to reduce the computing time required for this feature. 

Since the network simplex algorithm is based on spanning trees of the graph, we further compute a few features of a minimum spanning tree. The mean value, sum and standard deviation of the tree arc costs and capacities are calculated. Again, the mean value and the standard deviation are normalized by dividing by the maximum arc cost and capacity, respectively.

\begin{table}[H]
    \centering
    \begin{tabularx}{\textwidth}{|p{4.3cm}|X|}
    \hline
        \textbf{Feature} & \textbf{Measure}\\
        \hline
        number of vertices & count\\
        \hline
        number of arcs &  count\\
        \hline
        costs & max, min, mean, sum, standard deviation \\
        \hline
        capacity & max, min, mean, sum, standard deviation \\
        \hline
        supply & total supply, number supply/demand vertices\\
        \hline
        distance source sink & number of arcs\\
        \hline
        minimum spanning tree & mean, sum, standard deviation\\
        \hline
    \end{tabularx}
    \caption{Features calculated for each MCF instance.}
    \label{tab:features}
\end{table}

\subsection{ML algorithms}\label{subsec:mlalgorithms}
We use a bunch of different ML algorithms to solve our classification problem of predicting the fastest MCF algorithm for each instance. The classifiers come from the Python module Scikit-learn \cite{scikit-learn} and are based on the well-known algorithms k-nearest-neighbours, support vector machines, decision-trees and random forests, neural networks and AdaBoost. Each of these algorithms requires a set of hyperparameters as input. The performance of an ML algorithm can depend heavily on the set of hyperparameters used. Therefore, we perform a simple grid search to find suitable hyperparameters for each algorithm. In the following, we briefly describe each of the algorithms and the parameters we optimize during the grid search. For a more detailed description, we refer the reader to \cite{hastie2016elements}.\\
\textbf{k-nearest-neighbours classifier} saves all of the training data during the training process \cite{hastie2016elements}. It classifies by considering the $k$ nearest neighbours and performing a majority vote \cite{hastie2016elements}. As hyperparameters, it takes the number of neighbours $k$  and the type of weights. The type can either be uniform or distance. If the weights are set to uniform, during the majority vote every neighbour has the same weight. Otherwise, if the weights are set to distance, during the majority vote a neighbour which is closer to the instance to be classified has a higher weight than a neighbour which is further away. In both cases, the instance is classified by assigning it to the class with the highest sum of weights. During the grid search, the parameter $k$ is taken from the set $\{8, 10, 20, 50, 70, 90\}$ and the weight type is chosen as either uniform or distance.\\
\textbf{Support vector machine} finds seperating hyperplanes maximizing the distance between the hyperplane and training data \cite{hastie2016elements}. It can be thought of as a generalization of optimal seperating hyperplanes for the non-seperable case  \cite{hastie2016elements}. It takes as hyperparameters a regularization parameter $C$ and a kernel function. The kernel allows the SVM not to be a linear classifier. During training and classification, only scalar products are used. A non-linear kernel now represents a scalar product in a  higher dimensional space than the dimension the input data is provided in, in which the input data then is implicitly embedded. The parameter $C$ influences the penalty of points being near or on the wrong side of the hyperplane. During the grid search, $C$ is chosen from the set $\{0.5, 0.75, 1.0, 1.25, 1.5\}$ and the kernel function is chosen as either linear, polynomial or a radial basis function.\\
\textbf{Decision-trees} learn simple decision rules based on the features of the input data and, thus, partition the input space into different regions, each of which are then assigned to a class \cite{hastie2016elements}. The decision rules are to be traversed in a tree-like manner \cite{hastie2016elements}. The deeper the depth of the tree, the more possible decisions arise. This thus leads to a finer partition \cite{hastie2016elements}. This classifier can be parameterized in terms of its maximum depth - a strategy for splitting vertices during training - and the criterion function that measures the quality of each split. More precisely, during the grid search procedure, the maximum depth is chosen from the set $\{\infty, 3, 5, 8\}$. Further, we choose either the gini or the entropy function as the criterion, and we choose either the best or a random strategy as the splitting strategy. Additionally, another parameter which balances the weight of each class is either enabled or disabled.\\
\textbf{Random forest} is a learning method built on top of decision-trees \cite{hastie2016elements}. Multiple decision-trees are trained on random subsets of the features and subsets of the training data, and classification is done by simple majority voting \cite{hastie2016elements}. We optimize over the same hyperparameters as for decision-trees, except for the splitting strategy, since scikit-learn no longer provides this hyperparameter. The amount of decision-trees trained is also be parameterized. This number is chosen from the set $\{10, 50, 100, 200\}$.\\
\textbf{Multi-layer perceptron} is a feedforward neural network. It usually consists of multiple layers, each of which consist of a single linear transformation composed with a non-linear activation function \cite{hastie2016elements}. It can be parameterized in terms of the number of layers, number of neurons in each layer which determine the dimensionality of the linear transformation, the activation function, a learning rate and the optimization algorithm used. We optimize in terms of neural networks with either one layer and 100 neurons, two layers with 50 neurons each or 4 layers with 50, 100, 100 and 50 neurons. The activation function is chosen as either $tanh$ or $relu$, the learning rate is chosen from the set $\{0.001, 0.01, 0.1\}$ and the solver is either Stochastic Gradient Descent or ADAM.\\
\textbf{AdaBoost} starts by fitting a classifier and then fits copies of the classifier on data where weights of misclassified instances are adjusted \cite{hastie2016elements}. Classification is then done by a weighted majority vote \cite{hastie2016elements}. We parameterize over the number of trained classifiers taken from the set $\{5, 7, 9, 11, 13, 50\}$ and the learning rate taken from the set $\{0.8, 0.85, 1, 1.15, 1.3\}$.

\subsection{Training methodology}\label{subsec:training}
First, the entire dataset of 73130 feasible instances is split into a training and a test dataset, each consisting of 58504 and 14626 instances, respectively. All of the results presented in Section \ref{sec:results} are calculated using only the test dataset. The test dataset is not seen by any of the algorithms during training. For each algorithm described in Section \ref{subsec:mlalgorithms} a gridsearch over all possible hyperparameters is performed. For each combination of hyperparameters, we perform a 5-fold-cross-validation on the training dataset. The best set of hyperparameters in terms of mean accuracy within the cross validation is then saved. Finally, training is performed on the entire training dataset with the best hyperparameters. This approach is consistent with good practices in the literature \cite{hastie2016elements}, \cite{traintestsplit}.

\section{Results}\label{sec:results}
\begin{figure}[h]
 \centering
 \includegraphics[width=13cm,height=10cm,keepaspectratio]{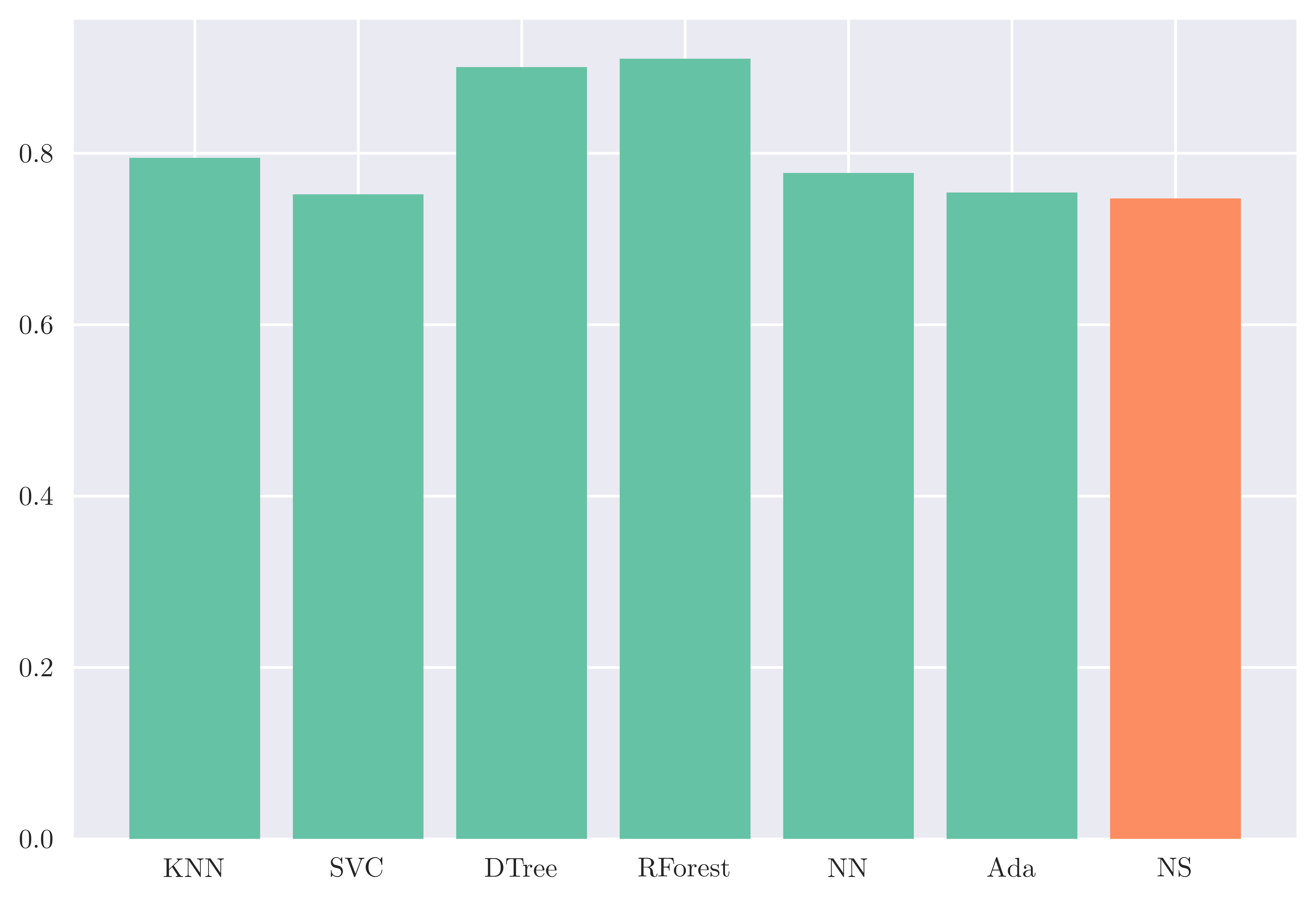}
 \caption{Accuracies of all classifiers in comparison to always choosing NS on 14626 test instances.}
 \label{fig:accuracies}
\end{figure}

In this section, we evaluate the accuracy of the classifiers described in Section~\ref{subsec:mlalgorithms} , i.e., we divide the number of correct predictions by the total number of predictions. Figure~\ref{fig:accuracies} shows the achieved accuracy of the six classifiers on the 14626 test instances.

The rightmost bar in Figure \ref{fig:accuracies} shows the accuracy of our baseline model, i.e., we always choose the network simplex algorithm to solve the instances. This would be an intuitive choice according to our observations in Subsection~\ref{subsec:data} and yields an accuracy of $74.72\%$. All the classifiers we trained achieve accuracies greater than $75\%$, with k-nearest-neighbours, decision tree and random forest achieving the best results. The k-nearest-neighbours classifier achieves an accuracy of almost $80\%$ with the hyperparameters of 20 neighbours and the weights set to distance. Decision tree and random forest each achieve an accuracy greater than $90\%$ with $90.05\%$ and $91.03\%$, respectively. For both classifiers, the criterion function entropy turns out to be the best and the different classes are not weighted. The decision tree classifier has a maximum depth of 8 and chooses the split strategy best. The maximum depth is not restricted for the random forest classifier and it considers the results of 200 decision trees. Table~\ref{tab:hyperpara} shows the hyperparameter settings that lead to the highest accuracies for all classifiers.

\begin{table}[h]
    \centering
    \begin{tabularx}{\textwidth}{|p{3cm}|X|}
        \hline
        \textbf{Classifier} & \textbf{Best hyperparameters} \\
        \hline
        KNN & neighbors: 20, weights: distance\\
        \hline
        SVM & C: 1.5, kernel: rbf\\
        \hline
        Decision-Tree & criterion: entropy, splitter: best, \newline max depth: 8, class weight: null\\
        \hline
        Random Forest & criterion: entropy, max depth: $\infty$, \newline estimators: 200, class weight: null\\
        \hline
        MLP & hidden layer sizes: 100, activation: tanh, \newline solver: adam, learning rate init: 0.001\\
        \hline
        AdaBoost & estimators: 5, learning rate: 0.85\\
        \hline
    \end{tabularx}
    \caption{Hyperparameters that yield the highest accuracy in the grid search for each ML classifier.}
    \label{tab:hyperpara}
\end{table}
In summary, among all the state of the art ML models we have studied, tree-based models seem to perform best in our particular setting of predicting the fastest algorithm for a given MCF instance. Thus, we can solve MCF problems, which have multiple applications in practice, in a more targeted and faster way by using a suitable algorithm.

\section{Conclusion}\label{sec:conclusion}
In this paper, we have reviewed seven state-of-the-art algorithms to solve the minimum cost flow problem. We created a representative dataset of 81,000 instances to evaluate the respective runtimes of these solvers for each of these instances. For this purpose, we used several random graph generators corresponding to the best practices known from the literature. We then formally defined the prediction task to be investigated, which is to predict the fastest minimum cost flow solver for any instance. To the best of our knowledge, we are the first to study this task for the minimum cost flow problem. It has turned out that the network simplex algorithm denotes the fastest in roughly 75\% of the cases. This intuitive assumption of always predicting the network simplex algorithm therefore served as a baseline model throughout this article. Further, we have explored various state-of-the-art machine learning models and trained them on the representative dataset to solve the above mentioned classification task. It has been shown that tree-based models performed particularly good at adapting and using the relevant structures in order to make appropriate predictions. After performing a gridsearch over a set of representative hyperparameters, the random forest classifier turned out to be the best classifier with an accuracy of over 91\% significantly outperforming the baseline model.

In the future, we would like to see this methodology applied to other combinatorial optimization problems of practical relevance. Certainly, there is plenty of room for further improvement by exploring other features, models and hyperparameters. Further, investigating how to automatically extract relevant features for graph-based combinatorial problems instead of using expert knowledge could be of particular interest. Another exciting research question would be to further investigate the split criteria of the tree-based models to better understand and explain the model results or the relevant structures of the minimum cost flow problem.

\section*{Acknowledgements}
This work was partially supported by the Carl Zeiss Foundation within the project \enquote{Ageing Smart -- Räume intelligent gestalten}.

\bibliographystyle{apalike}
\bibliography{biblio}

\begin{thebibliography}{}

\bibitem[Ahmady and Yeghaneh, 2022]{mcf_application_freight_traffic_1}
Ahmady, M. and Yeghaneh, Y.~E. (2022).
\newblock Optimizing the cargo flows in multi-modal freight transportation
  network under disruptions.
\newblock {\em Iranian Journal of Science and Technology, Transactions of Civil
  Engineering}, 46:453--472.

\bibitem[Bellman, 1958]{bellman1958routing}
Bellman, R. (1958).
\newblock On a routing problem.
\newblock {\em Quarterly of applied mathematics}, 16(1):87--90.

\bibitem[Bertsekas et~al., 1994]{generators}
Bertsekas, D.~P., Tseng, P., et~al. (1994).
\newblock Relax-iv: A faster version of the relax code for solving minimum cost
  flow problems.

\bibitem[Bokinge and Hasselstr{\"o}m, 1980]{mcf_application_transport_1}
Bokinge, U. and Hasselstr{\"o}m, D. (1980).
\newblock Improved vehicle scheduling in public transport through systematic
  changes in the time-table.
\newblock {\em European Journal of Operational Research}, 5(6):388--395.

\bibitem[Branco, 2011]{mcf_application_freight_traffic_4}
Branco, J. E.~H. (2011).
\newblock Estimating freight demand for north south railway.
\newblock {\em Journal of Transport Literature}, 5(4):17--50.

\bibitem[Busacker and Gowen, 1960]{ssp3}
Busacker, R.~G. and Gowen, P.~J. (1960).
\newblock A procedure for determining a family of minimum-cost network flow
  patterns.
\newblock Technical report, RESEARCH ANALYSIS CORP MCLEAN VA.

\bibitem[Dantzig, 2016]{ns1}
Dantzig, G. (2016).
\newblock {\em Linear programming and extensions}.
\newblock Princeton university press.

\bibitem[Dantzig, 1951]{ns2}
Dantzig, G.~B. (1951).
\newblock Application of the simplex method to a transportation problem.
\newblock {\em Activity analysis and production and allocation}.

\bibitem[Dezs{\H{o}} et~al., 2011]{lemon}
Dezs{\H{o}}, B., J{\"u}ttner, A., and Kov{\'a}cs, P. (2011).
\newblock {LEMON}--an open source {C}++ graph template library.
\newblock {\em Electronic Notes in Theoretical Computer Science},
  264(5):23--45.

\bibitem[Edmonds and Karp, 1972]{ssp4}
Edmonds, J. and Karp, R.~M. (1972).
\newblock Theoretical improvements in algorithmic efficiency for network flow
  problems.
\newblock {\em Journal of the ACM (JACM)}, 19(2):248--264.

\bibitem[Engels, 2011]{mcf_application_freight_traffic_3}
Engels, B. (2011).
\newblock {\em A Generalized Network Model for Freight Car Distribution}.
\newblock Dr. Hut Verlag M{\"u}nchen.

\bibitem[Ford~Jr and Fulkerson, 1962]{ford_and_fulkerson}
Ford~Jr, L. and Fulkerson, D. (1962).
\newblock {\em Flows in Networks, Princeton UniversityPress, Princeton, NJ,
  1962}.

\bibitem[Frangioni and Manca, 2006]{mcfalgorithms2}
Frangioni, A. and Manca, A. (2006).
\newblock A computational study of cost reoptimization for min-cost flow
  problems.
\newblock {\em INFORMS Journal on Computing}, 18(1):61--70.

\bibitem[F{\"u}genschuh et~al., 2006]{mcf_application_freight_traffic_5}
F{\"u}genschuh, A., Homfeld, H., Huck, A., and Martin, A. (2006).
\newblock Locomotive and wagon scheduling in freight transport.
\newblock In {\em 6th Workshop on Algorithmic Methods and Models for
  Optimization of Railways (ATMOS'06)}. Schloss Dagstuhl-Leibniz-Zentrum
  f{\"u}r Informatik.

\bibitem[Goldberg and Tarjan, 1988]{MMCC}
Goldberg, A. and Tarjan, R. (1988).
\newblock Finding minimum-cost circulations by canceling negative cycles.
\newblock In {\em Proceedings of the twentieth annual ACM symposium on Theory
  of computing}, pages 388--397.

\bibitem[Goldberg, 1997]{cs2}
Goldberg, A.~V. (1997).
\newblock An efficient implementation of a scaling minimum-cost flow algorithm.
\newblock {\em Journal of algorithms}, 22(1):1--29.

\bibitem[Hamacher and Tufekci, 1987]{mcf_application_evacuation_modeling_1}
Hamacher, H.~W. and Tufekci, S. (1987).
\newblock On the use of lexicographic min cost flows in evacuation modeling.
\newblock {\em Naval Research Logistics (NRL)}, 34(4):487--503.

\bibitem[Hassold and Ceder, 2014]{mcf_application_transport_3}
Hassold, S. and Ceder, A.~A. (2014).
\newblock Public transport vehicle scheduling featuring multiple vehicle types.
\newblock {\em Transportation Research Part B: Methodological}, 67:129--143.

\bibitem[Hastie et~al., 2016]{hastie2016elements}
Hastie, T., Tibshirani, R., and Friedman, J. (2016).
\newblock {\em The Elements of Statistical Learning: Data Mining, Inference,
  and Prediction}.
\newblock Springer series in statistics. Springer.

\bibitem[Iri, 1960]{ssp1}
Iri, M. (1960).
\newblock A new method of solving transportation-network problems.
\newblock {\em Journal of the Operations Research Society of Japan}, 3(1):2.

\bibitem[Jewell, 1958]{ssp2}
Jewell, W.~S. (1958).
\newblock Optimal flow through networks.
\newblock In {\em Operations Research}, volume~6, pages 633--633. INST
  OPERATIONS RESEARCH MANAGEMENT SCIENCES 901 ELKRIDGE LANDING RD.

\bibitem[Kim et~al., 2016]{mcf_application_freight_traffic_2}
Kim, N.~S., Park, B., and Lee, K.-D. (2016).
\newblock A knowledge based freight management decision support system
  incorporating economies of scale: multimodal minimum cost flow optimization
  approach.
\newblock {\em Information Technology and Management}, 17(1):81--94.

\bibitem[Klein, 1967]{scc}
Klein, M. (1967).
\newblock A primal method for minimal cost flows with applications to the
  assignment and transportation problems.
\newblock {\em Management Science}, 14(3):205--220.

\bibitem[Korte and Vygen, 2008]{theory}
Korte, B. and Vygen, J. (2008).
\newblock Combinatorial optimization: Theory and algorithms.
\newblock {\em Springer, Third Edition, 2005.}

\bibitem[Kovács, 2015]{mcfalgorithms}
Kovács, P. (2015).
\newblock Minimum-cost flow algorithms: An experimental evaluation.
\newblock {\em Optimization Methods and Software}, 30(1):94--127.

\bibitem[Kuhn, 1955]{kuhn}
Kuhn, H.~W. (1955).
\newblock The hungarian method for the assignment problem.
\newblock {\em Naval research logistics quarterly}, 2(1-2):83--97.

\bibitem[Musliu and Schwengerer, 2013]{gcpfeature}
Musliu, N. and Schwengerer, M. (2013).
\newblock Algorithm selection for the graph coloring problem.
\newblock In {\em International conference on learning and intelligent
  optimization}, pages 389--403. Springer.

\bibitem[Pedregosa et~al., 2011]{scikit-learn}
Pedregosa, F., Varoquaux, G., Gramfort, A., Michel, V., Thirion, B., Grisel,
  O., Blondel, M., Prettenhofer, P., Weiss, R., Dubourg, V., Vanderplas, J.,
  Passos, A., Cournapeau, D., Brucher, M., Perrot, M., and Duchesnay, E.
  (2011).
\newblock Scikit-learn: Machine learning in {P}ython.
\newblock {\em Journal of Machine Learning Research}, 12:2825--2830.

\bibitem[Pihera and Musliu, 2014]{tspfeature}
Pihera, J. and Musliu, N. (2014).
\newblock {Application of Machine Learning of Algorithm Selection for TSP}.
\newblock In {\em 2014 IEEE 26th International Conference on Tools with
  Artificial Intelligence}, pages 47--54.

\bibitem[Pyakurel and Dhamala, 2014]{mcf_application_evacuation_modeling_2}
Pyakurel, U. and Dhamala, T.~N. (2014).
\newblock Earliest arrival contraflow model for evacuation planning.
\newblock {\em Neural, Parallel, and Scientific Computations}, 22:287--294.

\bibitem[Pyakurel and Dhamala, 2015]{mcf_application_evacuation_modeling_3}
Pyakurel, U. and Dhamala, T.~N. (2015).
\newblock Models and algorithms on contraflow evacuation planning network
  problems.
\newblock {\em International Journal of Operations Research}, 12(2):36--46.

\bibitem[Quanrud, 2018]{mcf_application_transport_2}
Quanrud, K. (2018).
\newblock Approximating optimal transport with linear programs.
\newblock {\em arXiv preprint arXiv:1810.05957}.

\bibitem[Rice, 1976]{algselection}
Rice, J.~R. (1976).
\newblock The algorithm selection problem.
\newblock In {\em Advances in computers}, volume~15, pages 65--118. Elsevier.

\bibitem[R{\"o}ck, 1980]{cos}
R{\"o}ck, H. (1980).
\newblock Scaling techniques for minimal cost network flows.
\newblock {\em Discrete structures and algorithms}.

\bibitem[Sifaleras, 2013]{mcfsurvey}
Sifaleras, A. (2013).
\newblock {Minimum cost network flows: Problems, algorithms, and software}.
\newblock {\em Yugoslav journal of operations research}, 23:3--17.

\bibitem[Tardos, 1985]{tardos}
Tardos, {\'E}. (1985).
\newblock A strongly polynomial minimum cost circulation algorithm.
\newblock {\em Combinatorica}, 5(3):247--255.

\bibitem[Wang and Wang, 2019]{mcf_application_evacuation_modeling_4}
Wang, Y. and Wang, J. (2019).
\newblock Integrated reconfiguration of both supply and demand for evacuation
  planning.
\newblock {\em Transportation Research Part E: Logistics and Transportation
  Review}, 130:82--94.

\bibitem[Xu and Goodacre, 2018]{traintestsplit}
Xu, Y. and Goodacre, R. (2018).
\newblock On splitting training and validation set: A comparative study of
  cross-validation, bootstrap and systematic sampling for estimating the
  generalization performance of supervised learning.
\newblock {\em Journal of Analysis and Testing}, 2.

\bibitem[Yamada, 1996]{mcf_application_evacuation_modeling_5}
Yamada, T. (1996).
\newblock A network flow approach to a city emergency evacuation planning.
\newblock {\em International Journal of Systems Science}, 27(10):931--936.

\end{thebibliography}

%
%
%
\end{document}